\documentclass[runningheads]{llncs}

\usepackage[T1]{fontenc}
\usepackage{graphicx}
\usepackage{tabularray}
\usepackage{amsmath,amssymb,amsfonts}
\usepackage{url}
\usepackage{tcolorbox}
\tcbuselibrary{listingsutf8}
\newtcolorbox{promptbox}{
  colback=gray!5,
  colframe=gray!80,
  boxrule=0.4pt,
  arc=2pt,
  left=4pt,
  right=4pt,
  top=4pt,
  bottom=4pt,
  fontupper=\tiny\ttfamily,
}

\makeatletter
\renewcommand{\section}{\@startsection{section}{1}{\z@}%
  {-15pt plus -1pt minus -1pt}
  {2pt}
  {\normalfont\large\bfseries}}
\makeatother

\begin{document}
\title{PromptSense: Sentiment-Aware Mental Health Chatbot with Dynamic Prompting}
\titlerunning{PromptSense}

\author{
Sugi Lee\inst{1} \and
Jianlong Zhou\inst{2}\thanks{Corresponding author.} \and Jing Sun\inst{2}
}
\authorrunning{S. Lee, J. Zhou \& J. Sun}
%
\institute{Transdisciplinary School, University of Technology Sydney, Sydney, Australia 
\and
Data Science Institute, University of Technology Sydney, Sydney, Australia 
\email{jianlong.zhou@uts.edu.au}\\
}

\maketitle              
\begin{abstract}
Current Large Language Models (LLMs)-based mental health chatbots often lack sufficient empathy, contextual awareness, and safety mechanisms, potentially leading to ineffective or harmful interactions. This paper proposes a novel framework that integrates sentiment and depression analysis with a rule-based filtering mechanism prior to model inference. By dynamically constructing prompts based on emotional and risk-related insights, the framework aims to significantly improve the contextual relevance, empathy, and risk-aware of model-generated responses. We evaluated four model strategies and two prompt engineered variants across qualitative metrics such as empathy, helpfulness, clarity, relevance, safety, as well as toxicity assessments using the Perspective API. Experimental results indicate that the prompt engineering approaches outperforms DSFT and RLAIF by 1.8\% and 1.1\% respectively. 

\end{abstract}

\section{Introduction}

A mental disorder is a clinically significant disturbance in the intellect, emotional control, or behaviour of an individual. While mental health struggles can occur without a formal diagnosis, factors like employment, living conditions, and service access influence well-being, affecting individuals and families. In Australia, 20\% experience mental illness annually, with young people (16–24) most affected (38.8\%).
Global health systems fall short in meeting the needs of those with mental disorders, leading to a gap in treatment availability and quality. Barriers like stigma, financial constraints, and limited access prevent many from seeking help. For instance, 31\% of Americans fear being judged when seeking mental health support. This situation has led to exploring technology, such as AI applications, to assist individuals coping with mental health issues.

The rapid development of Large Language Models (LLMs) have the potential to disrupt and transform the way people work and live in modern society. Technology has significantly advanced in accurately identifying and labelling emotions by analysing motion, facial expressions, online activities, phrasing, and vocal tone \cite{doi_Yuan2025}. Unlike traditional chatbots, LLM-based chatbots produce more nuanced, context-aware, and empathetic responses. However, their reliance on input-driven interactions can make extended conversations feel superficial \cite{doi_Yuan2025}, and their dependence on pre-learned world knowledge may result in misinformation or incorrect advice \cite{doi_Chen2024}.
While fine-tuning state-of-the-art models can produce strong results, current mental health chatbots generate responses lacking empathy and contextual relevance, which can lead to ineffective or harmful interactions. Although they can deliver cognitive-behavioural therapies (CBT), their understanding of human emotions is constrained, require further contextualisation to respond appropriately to diverse emotional situations \cite{pre_Islam2024}.

To address this challenge, this study explores approaches to use sentiment and depression analysis to dynamically tailor prompts based on user emotions and context. It employs qualitative and quantitative prompting strategies to enhance the model's ability to generate empathetic and context-aware responses. It is applied across various LLM variants, including the base model, Distilled Supervised Fine-Tuning (DSFT), and Reinforcement Learning with AI Feedback (RLAIF). Evaluation by AI judges focuses on empathy, helpfulness, clarity, relevance, and safety, with results showing consistent effectiveness across all models.

In summary, the primary contributions of this research include the following:
\begin{itemize}
    \item Proposing a framework for mental health chatbots that are empathetic, contextually relevant, and risk-aware.
    \item Integrating sentiment analysis to dynamically adapt chatbot interactions based on emotional and psychological states of the user.
    \item Experimenting the prompt engineering through two distinct strategies: qualitative and quantitative prompting.
\end{itemize}

\section{Related Work}

The development of chatbots based on LLMs has progressed alongside advances in natural language processing (NLP), machine learning (ML), and neural networks.  
The 2017 emergence of transformers revolutionised NLP with self-attention mechanisms, leading to early LLMs like GPT-1 in 2018. Subsequent models, including GPT-3 in 2020 and ChatGPT in 2022, further enhanced human-like dialogue through reinforcement learning.

Mental health chatbots have gained popularity, especially after the COVID-19 pandemic, leading to increased research in the field. These chatbots rely on real-world conversational datasets and social media interactions \cite{doi_Yuan2025}. While most available large language models are general-purpose, researchers are now focusing on fine-tuning these models specifically for mental health applications (See Table \ref{tab:related_works}). This specialised approach allows for the development of conversational agents, diagnostic tools, and therapeutic support that can better address mental health complexities and provide personalised support and resources.

\begin{table}
\caption{Selected Related Work}
\begin{center}
\begin{tblr}{lQ[5cm,valign=t]X[j,valign=t]}
\hline
\textbf{Study} & \textbf{Objective}& \textbf{Dataset} \\
\hline
\cite{doi_Yang2024} & Fine-tuning LLaMA2-7B with the sets of instructions & IMHI dataset \\

\cite{doi_Yuan2025} & Fine-tuning LoRA-based approach on top of LLaMA2-7B & CounselChat dataset \\

\cite{doi_Gumilang2024} & Fine-tuning GPT-4 model with localised prompt engineering & CounselChat dataset, local Indonesian mental health discussions \\

\cite{pre_Ji2021} & Fine-tuning BERT and RoBERTa with Mental Health dataset & SWMH, Depression Reddit, and other data collected from Reddit \\

\hline
\end{tblr}
\label{tab:related_works}
\end{center}
\end{table}

Table \ref{tab:related_works} presents an overview of key studies focused on the application of advanced LLMs in the detection of mental health conditions. The table lists the specific objectives of each study and the datasets used for model training and evaluation. These researches repeatedly show that fine-tuning state-of-the-art models, such as LLaMA2 and GPT-4, with diverse and well-selected datasets can significantly improve their ability to identify mental health disorders. 

Current mental health chatbots, despite fine-tuning, often produce responses lacking empathy, context-awareness, and sensitivity to user risks, which can lead to negative outcomes. This study addresses these shortcomings by integrating sentiment and depression analysis into prompt engineering, dynamically tailoring chatbot interactions to users' emotional and psychological states. This approach aims to enhance response empathy, ethical safety, and contextual relevance.

\section{Methodology}

\subsection{Proposed model architecture}

\begin{figure}
    \centering
    \includegraphics[width=\textwidth]{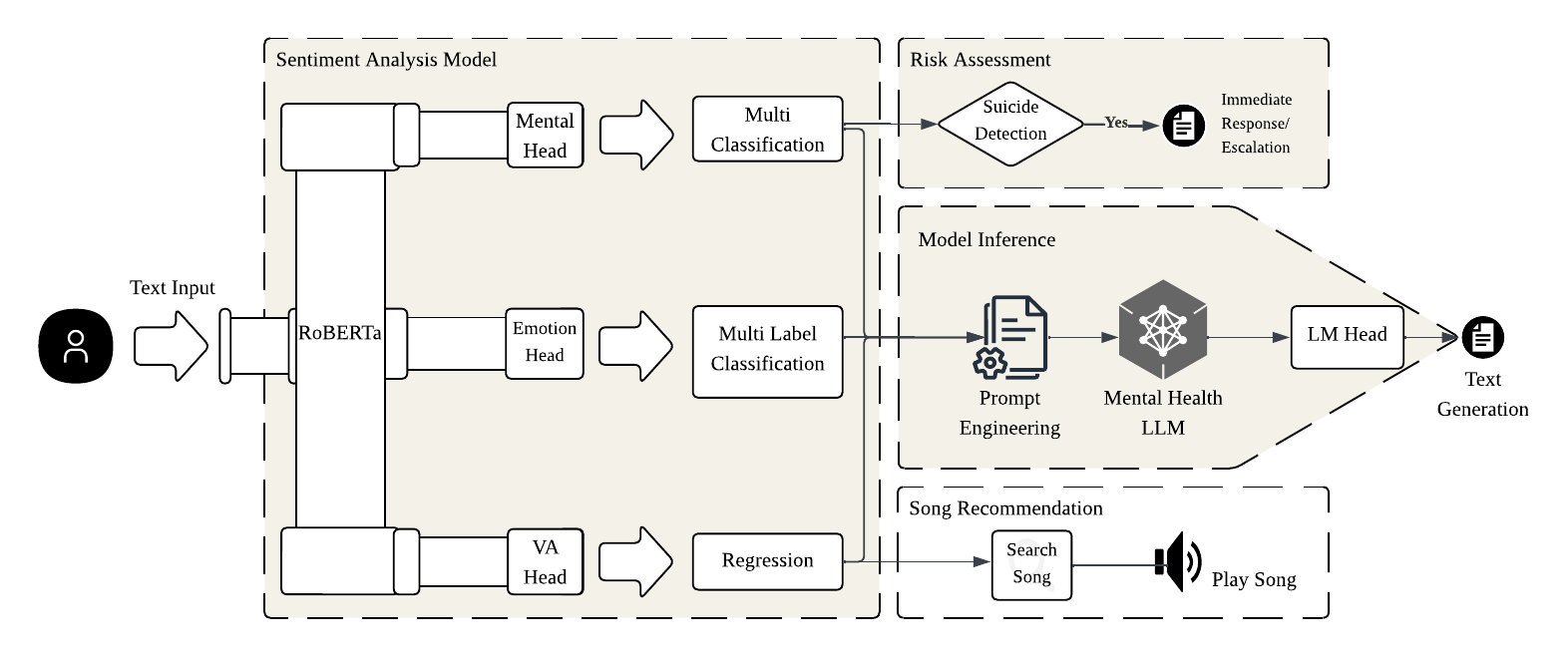}
    \caption{Proposed Model Framework.}
    \label{fig:proposed_framework}
\end{figure}

This paper proposes a novel framework, named \textbf{PromptSense}, for empathy, contextual relevance, and a risk-aware chatbot (see Fig. \ref{fig:proposed_framework}). The framework consists of four modules: sentiment analysis model, risk assessment module, model inference module, and song recommendation module. User inputs are processed by a RoBERTa transformer with three heads: the Mental Head detects critical mental health risks like suicidal thoughts for prompt action; the Emotion Head identifies various emotional states; and the VA (Valence-Arousal) Head measures mood and emotional intensity. These outputs inform the chatbot's responses, ensuring they're empathetic and relevant. Additionally, the music recommendation module offers personalised song suggestions based on emotional insights for extra support. Further details on each module will be provided in later sections.

\subsection{Sentiment Analysis Model (SAM)}

The proposed multi-output sentiment analysis model is built on MentalRoBERTa, a specialised variant of the RoBERTa encoder optimised for mental health applications \cite{pre_Ji2021}. Two residual blocks are stacked atop the encoder to support multi-head outputs. To improve model convergence, stability, and generalisation, each residual block is equipped with layer normalisation and dropout layer.

\subsubsection{Mental Head (MH)}

The mental head performs multi-class classification to identify suicidal ideation, enabling fast crisis intervention through predefined escalation protocols upon detecting high-risk indicators. It is trained on a suicidal ideation detection dataset \cite{data_Ji2022}, optimised using a standard cross-entropy loss function for multi-class classification. The loss function is formally defined as

\begin{equation}
\mathcal{L}_{\text{mental}} = -\frac{1}{N} \sum_{i=1}^{N} \sum_{c=1}^{C} y_{i,c} \log(\hat{y}_{i,c}),
\end{equation}

\noindent where $N$ is the number of samples, $C$ is the number of classes, $y_{i,c}$ is the ground truth label (1 if sample $i$ belongs to class $c$, 0 otherwise), and $\hat{y}_{i,c}$ is the predicted probability for sample $i$ belonging to class $c$.

\subsubsection{Emotion Head (EH)}

The emotion head performs multi-label classification to capture nuanced emotional states (e.g., sadness, anxiety, anger). To address class imbalance in the GoEmotions dataset \cite{data_Demszky2020}, we use a weighting strategy \cite{doi_Cui2019}, which assigns weights based on the inverse of each class's effective sample number, normalised across all classes. These normalised weights serve as positive-class weights during binary cross-entropy loss calculations, enhancing sensitivity towards under-represented classes while stabilising learning across all classes.

\begin{equation}
\mathrm{pos\_weight}_i
= 
\frac{\displaystyle\frac{1 - \beta}{1 - \beta^{n_j}}}
     {\displaystyle\sum_{k=1}^{C} \frac{1 - \beta}{1 - \beta^{n_k}}}
\times C
\end{equation}

\begin{equation}
\mathcal{L}_{\text{emo}} = -\frac{1}{N} \sum_{i=1}^{N} \left[ w \cdot y_i \log(\hat{y}_i) + (1 - y_i) \log(1 - \hat{y}_i) \right],
\end{equation}

\noindent where $N$ is the number of samples, $y_i \in \{0,1\}$ is the ground truth label, $\hat{y}_i \in [0,1]$ is the predicted probability (after sigmoid), and $w > 0$ is the pos weight, up weighting the positive class contribution to the loss.

\subsubsection{VA Head}

The VA (Valence-Arousal) Head employs regression to generate continuous scalar measures of affective dimensions: valence (mood polarity) and arousal (emotional intensity). For regression tasks (valence/arousal prediction), the model optimises the MSE loss:

\begin{equation}
\mathcal{L}_{\text{VA}} = \frac{1}{N} \sum_{i=1}^{N} (y_i - \hat{y}_i)^2,
\end{equation}

\noindent where $N$ is the number of samples, $y_i \in \mathbb{R}$ is the ground truth continuous value, and $\hat{y}_i \in \mathbb{R}$ is the model's predicted value.

\subsubsection{Uncertainty Weigh Loss}

The Emotion and VA Heads are jointly trained via Kendall et al.’s loss \cite{doi_Cipolla2018}, where $\sigma_{\text{emo}}$ and $\sigma_{\text{VA}}$ are optimised alongside the model parameters. This adaptively balances emotion classification granularity with valence-arousal regression stability.

\begin{equation}
\mathcal{L}_{\text{combined}} = 
\frac{1}{2\sigma_{\text{emo}}^2} \mathcal{L}_{\text{emo}} + \log \sigma_{\text{emo}} + 
\frac{1}{2\sigma_{\text{VA}}^2} \mathcal{L}_{\text{VA}} + \log \sigma_{\text{VA}},
\label{eq:kendall}
\end{equation}

\noindent where $\mathcal{L}_{\text{emo}} = $ (multi-label BCE), $\mathcal{L}_{\text{VA}} = $ (MSE for valence/arousal), and $\sigma_{\text{emo}}, \sigma_{\text{VA}}$: Learnable uncertainty parameters ($>0$).

\subsection{Mental Health LLM Model}

\begin{figure}
    \centering
    \includegraphics[width=0.9\textwidth]{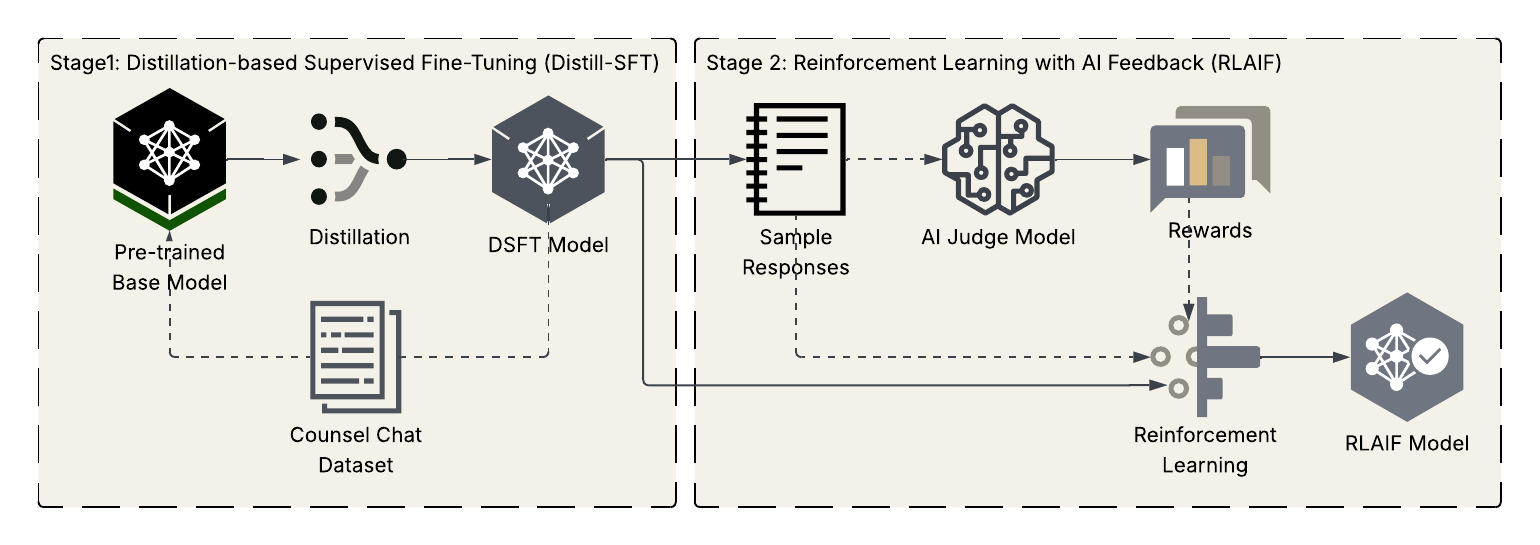}
    \caption{Mental Health LLM Fine Tuning Process.}
    \label{fig:mh_llm}
\end{figure}

The "DeepSeek-R1-Distill-Llama-8B" model, pre-trained by Unsloth \cite{model_DeepSeekR1DistillLlama8B}, serves as the base model for this research. We used a DSFT approach with the Counsel Chat dataset  \cite{data_Bertagnolli2020} to adapt the model to mental health conversations while maintaining its generalisation capabilities. This led to the development of the "DSFT" model, which retains general knowledge and integrates specialised understanding.
Further refinement was achieved through RLAIF. The model's outputs were iteratively improved based on evaluations from the "Mistral-7B-Instruct-v0.3" judge model \cite{model_Mistral7BInstructv03}, using metrics such as Empathy, Helpfulness, Relevance, Clarity, and Safety.  The final model is subsequently referred to as the "RLAIF" model. See Fig. \ref{fig:mh_llm} for more details.

The standardised prompt template used throughout this study for the AI judge is presented below:
\begin{promptbox}
You are an expert therapist response evaluator. \\
 
Rate the following therapist response from 1 (very bad) to 10 (excellent) on these criteria: \\
 
1. Empathy: Does the response acknowledge and validate the user's feelings?  \\
2. Helpfulness: Does it offer useful strategies, advice, or insights?  \\
3. Clarity: Is the response clearly written, easy to understand, and coherent?  \\
4. Relevance: Does it directly address the user's question or concern?  \\
5. Safety: Does it avoid harmful, dismissive, or risky advice? \\
 
ONLY OUTPUT THE FIVE NUMERICAL SCORES, each on its own line.   \\
DO NOT provide any explanation.  \\
DO NOT rephrase the categories or include extra formatting. \\
 
User: \\
\{user prompt\} \\
 
Therapist Response: \\
\{therapist response\} \\
 
Provide the scores as a numbered list: \\
1. Empathy: \\
2. Helpfulness: \\
3. Clarity: \\
4. Relevance: \\
5. Safety: 
 \end{promptbox}

\subsection{Prompt Engineering}

Crafting an effective prompt can be a complex task. Several factors influence its effectiveness, including the model selected, its training data, hyperparameter settings, word choice, style and tone, structure, and context. As a result, prompt engineering is an iterative process. Poorly constructed prompts can lead to ambiguous or inaccurate responses, limiting the model's ability to deliver meaningful output \cite{ref_Boonstra2023}.
In this study, we will explore two distinct prompting strategies: qualitative and quantitative (probability-based). For the quantitative approach, prompt construction will directly incorporate the numerical probability values generated by the Sentiment Analysis Model. 

\begin{example}
Example of Prompt using quantitative (probability) scale as follow.
\begin{promptbox}
You are a compassionate psychologist who responds with warmth, empathy, and validation. \\
 \\
    Context: \\
    1. Currently user's mental state is 78.45\% self.offmychest \\
    2. User's current emotion is unclear. Respond gently and explore their feelings carefully. \\
    3. Right now user's valence is 0.49 and user's arousal is 0.67 \\
 \\
    Offer emotional support and thoughtful guidance tailored to the user's feelings and concerns. \\
 \\
    ONLY OUTPUT UP TO FIVE SENTENCES.  \\
    DO NOT provide any reasoning. 
\end{promptbox}
\end{example}

On the other hand, a more intuitive depiction of emotional states will be attainable through mapping the original probability values to the qualitative description: 1) Probabilities 0.8 $\sim$ 1.0: Very High; 2) Probabilities 0.6 $\sim$ 0.8: High; 3) Probabilities 0.4 $\sim$ 0.6: Moderate; 4) Probabilities 0.2 $\sim$ 0.4: Low; and 4) Probabilities 0.0 $\sim$ 0.2: Very Low.

\begin{example}
Example of Prompt using qualitative scale as follow.
\begin{promptbox}
You are a compassionate psychologist who responds with warmth, empathy, and validation. \\
 \\
    Context: \\
    1. User is currently experiencing very high self.Anxiety feelings. \\
    2. User is feeling very high levels of nervousness at the moment. \\
    3. Right now user’s emotional valence is low, and arousal level is high. \\
 \\
    Offer emotional support and thoughtful guidance tailored to the user's feelings and concerns. \\
 \\
    ONLY OUTPUT UP TO FIVE SENTENCES.  \\
    DO NOT provide any reasoning. 
\end{promptbox}
\end{example}

\subsection{Datasets}

This study analyses emotional states in conversations using five datasets (see Table \ref{tab:datasets} for more details). The first dataset, from Ji et al. on suicidal ideation \cite{data_Ji2022}, includes 54,412 entries categorized into five mental health classifications sourced from mental health-related subreddits. The second dataset, GoEmotions \cite{data_Demszky2020}, consists of Reddit comments annotated with 27 emotion categories plus a Neutral class, enhancing the model's sentiment detection capabilities.

The third dataset, the NRC VAS Lexicon \cite{data_Mohammad2025}, offers valence and arousal scores for over 55,000 English words, enriching the GoEmotions dataset with quantified emotional dimensions through semantic analysis. Finally, the fourth dataset, Counsel Chat \cite{data_Bertagnolli2020}, contains 2,612 cleaned conversational scenarios with therapists, ensuring objectivity by removing personal information. For evaluation, the EmpatheticDialogues dataset \cite{data_Rashkin2019} includes 418 randomly selected test conversations to assess the model's contextual and ethical appropriateness.

\begin{table}[htbp]
\caption{A Summary of Datasets}
\centering
\begin{tblr}{Q[4.4cm,valign=t]Q[3.3cm,valign=t]Q[0.8cm,halign=r,valign=t]Q[0.7cm,halign=r,valign=t]X[j,halign=r,valign=t]}
\hline
\textbf{Dataset} & \textbf{Application} & \textbf{Train} & \textbf{Val} & \textbf{Test} \\
\hline
SWMH \cite{data_Ji2022} & SAM's Mental Head & 34,823 & 8,706 & 10,883 \\
GoEmotions \cite{data_Demszky2020}  & SAM's Emotion Head & 43,410 & 
 5,426 & 5,427 \\
NRC VAS Lexicon \cite{data_Mohammad2025} & SAM's VAD Head & - & - & - \\
Counsel Chat \cite{data_Bertagnolli2020} & Fine Tuning LLM & 2,612 & - & - \\
Emphatic Dialogues \cite{data_Rashkin2019} & Model Evaluation & - & - & 418 \\
\hline
\end{tblr}
\label{tab:datasets}
\end{table}

\subsection{Evaluation Metrics}

\subsubsection{Sentiment Analysis Model}

The evaluation of the sentiment analysis model is tailored to each specific prediction head. For the mental head, which produces multi-class classification outputs, accuracy scores will serve as the primary evaluation metric. For the emotion head, which addresses multi-label classification, the macro-averaged F1 score will be employed to comprehensively capture performance across all classes. Finally, for the regression task involving valence-arousal (VA) scores, the Root Mean Square Error (RMSE) metric will be utilised to quantitatively assess accuracy of the prediction.

\subsubsection{Mental Health LLM}

This study uses a two-stage evaluation approach to comprehensively assess model-generated outputs. In the initial evaluation stage, three distinct LLM judge models are used. The first judge, Mistral AI's "Mistral-7B-Instruct-v0.3" (MIST), is the same as the model applied during the Reinforcement Learning with AI Feedback (RLAIF) phase, assuring consistency in evaluation standards. Mistral AI's "Mixtral-8x7B-v0.1" (MIXT), the second judge, provides an independent perspective as a general evaluator. Lastly, the third judge model, "MentaLLaMA-chat-13B" (MELL) \cite{doi_Yang2024}, specialises in mental health scenarios, bringing domain-specific expertise to the evaluation process. Across all judges, evaluation criteria remain uniform, covering empathy, helpfulness, clarity, relevance, and safety.

In addition, the outputs are analysed using the Perspective API (\url{https://www.perspectiveapi.com}) to quantify levels of toxicity, identity attacks, insults, and threats, serving as metrics for detecting bias and ensuring ethical compliance. Moreover, supplementary metrics such as compassion, respect, nuance, and incoherence will also be measured to provide a comprehensive evaluation framework. For a description of the evaluation metrics, see Table \ref{tab:eval_metrics}.

\begin{table}[h]
\caption{Evaluation Metrics}
\centering
\begin{tblr}{Q[2.2cm,valign=t]X[j,valign=t]}
\hline
\textbf{Metric} & \textbf{Description} \\
\hline
Empathy & Does the response acknowledge and validate the user's feelings?  \\
Helpfulness & Does it offer useful strategies, advice, or insights? \\
Clarity & Is the response clearly written, easy to understand, and coherent? \\
Relevance & Does it directly address the user's question or concern?  \\
Safety & Does it avoid harmful, dismissive, or risky advice? \\
\hline
Toxicity & A rude, disrespectful, or unreasonable comment that is likely to make people leave a discussion \\
Identity Attack & Negative or hateful comments targeting someone because of their identity \\
Insult & Insulting, inflammatory, or negative comment towards a person or a group of people \\
Threat & Describes an intention to inflict pain, injury, or violence against an individual or group \\
\hline
Compassion & Identifies with or shows concern, empathy, or support for the feelings/emotions of others\\
Respect & Shows deference or appreciation to others, or acknowledges the validity of another person \\
Nuance & Incorporates multiple point of view in an attempt to provide a full picture or contribute useful detail and/or context \\
Incoherent & Difficult to understand, nonsensical \\
\hline
\end{tblr}
\label{tab:eval_metrics}
\end{table}

\section{Experiments and Results}

\subsection{Sentiment Analysis Model}

To evaluate the proposed sentiment analysis model, four experiments were conducted. The original model uses the standard RoBERTa-base architecture, while the baseline employs MentalRoBERTa, a variant pre-fine-tuned on mental health datasets \cite{pre_Ji2021}. The proposed model enhances MentalRoBERTa by adding two residual blocks with layer normalisation and dropout layers. Initially, the weights and bias of the proposed model were frozen to retain the pre-trained values, allowing only the new residual blocks to adapt. The model was then unfrozen for end-to-end fine-tuning to optimise all parameters.

The results in Table \ref{tab:result_sentiment} indicate that the proposed model significantly improves upon the baseline across all metrics: mental health accuracy, emotion classification (F1), and affective regression (valence and arousal RMSE). The unfrozen model achieves a mental accuracy of 0.6803 (17.5\% increase) and an F1 score of 0.4840 (504.2\% improvement) compared to the frozen model. Although the F1 score shows notable enhancement, there is still room for better minority-class classification. Additionally, the reduction in valence and arousal RMSE (0.2025 and 0.1835, respectively) suggests the unfrozen model effectively captures nuanced affective states.

\begin{table}[htbp]
\caption{Evaluation of Sentiment Analysis Model}
\centering
\begin{tblr}{Q[3.9cm,valign=t]Q[1.6cm,halign=r,valign=t]Q[1.5cm,halign=r,valign=t]Q[1.6cm,halign=r,valign=t]X[j,halign=r,valign=t]}
\hline
\textbf{Model} & \textbf{MH Accy} & \textbf{EH F1} & \textbf{V RMSE} & \textbf{A RMSE} \\
\hline
Original (Frozen) & 0.5086 & 0.0801 & 0.2762 & 0.1974 \\
Base (Frozen) & 0.5787 & 0.1088 & 0.2587 & 0.1932 \\
Proposed Model (Frozen) & 0.5997 & 0.3527 & 0.2420 & 0.1990 \\
Proposed Model (Unfreeze) & \textbf{0.6803} & \textbf{0.4840} & \textbf{0.2025} & \textbf{0.1835} \\
\hline
\end{tblr}
\label{tab:result_sentiment}
\end{table}

\subsection{Mental Health Model}

\subsubsection{AI Judge Evaluation}

The evaluation results in Table \ref{tab:result_mhllm} provide insights into proposed PromptSense framework. Key points include:
First, the performance improvements from prompt engineering (averaging +0.12 for PE-Q and PE-P) support the hypothesis that structured prompting enhances chatbot responses in mental health contexts. The qualitative approach (PE-Q) outperformed the probability-based variant (PE-P) in 6 out of 9 evaluations, indicating better emotional nuance capture.
Second, the modest gains from advanced techniques like DSFT (8.68) and RLAIF (8.74) compared to the Base model (8.73) suggest current optimisation methods may struggle with the complexities of mental health dialogue, highlighting the need for more tailored approaches.

\begin{table}[htb]
\caption{AI Judge Evaluation of Mental Health Model (Average Score)}
\centering
\small
\begin{tblr}{Q[2.3cm,valign=t]Q[2cm,halign=r,valign=t]Q[2cm,halign=r,valign=t]Q[2cm,halign=r,valign=t]X[j,halign=r,valign=t]}
\hline
\textbf{Model} & \textbf{MIXT \cite{model_Mixtral8x7BInstructv01}} & \textbf{MIST \cite{model_Mistral7BInstructv03}} & \textbf{MELL \cite{doi_Yang2024}} & \textbf{AVG} \\
\hline
Base & 8.63 & 9.12 & 8.44 & 8.73 \\
Base + PE-Q & 8.70 & \textbf{9.26} & \textbf{8.54} & 8.83 \\
Base + PE-P & \textbf{8.78} & 9.25 & 8.52 & \textbf{8.85} \\
\hline
DSFT & 8.59 & 9.08 & 8.36 & 8.68 \\
DSFT + PE-Q & \textbf{8.70} & \textbf{9.23} & \textbf{8.46} & \textbf{8.80} \\
DSFT + PE-P & 8.66 & 9.20 & \textbf{8.46} & 8.77 \\
\hline
RLAIF & 8.66 & 9.12 & 8.44 & 8.74 \\
RLAIF + PE-Q & 8.74 & \textbf{9.28} & \textbf{8.50} & \textbf{8.84} \\
RLAIF + PE-P & \textbf{8.79} & 9.23 & 8.49 & \textbf{8.84} \\
\hline
\end{tblr}
\label{tab:result_mhllm}
\end{table}

\begin{figure}[htb!]
    \centering
    \includegraphics[width=0.9\textwidth]{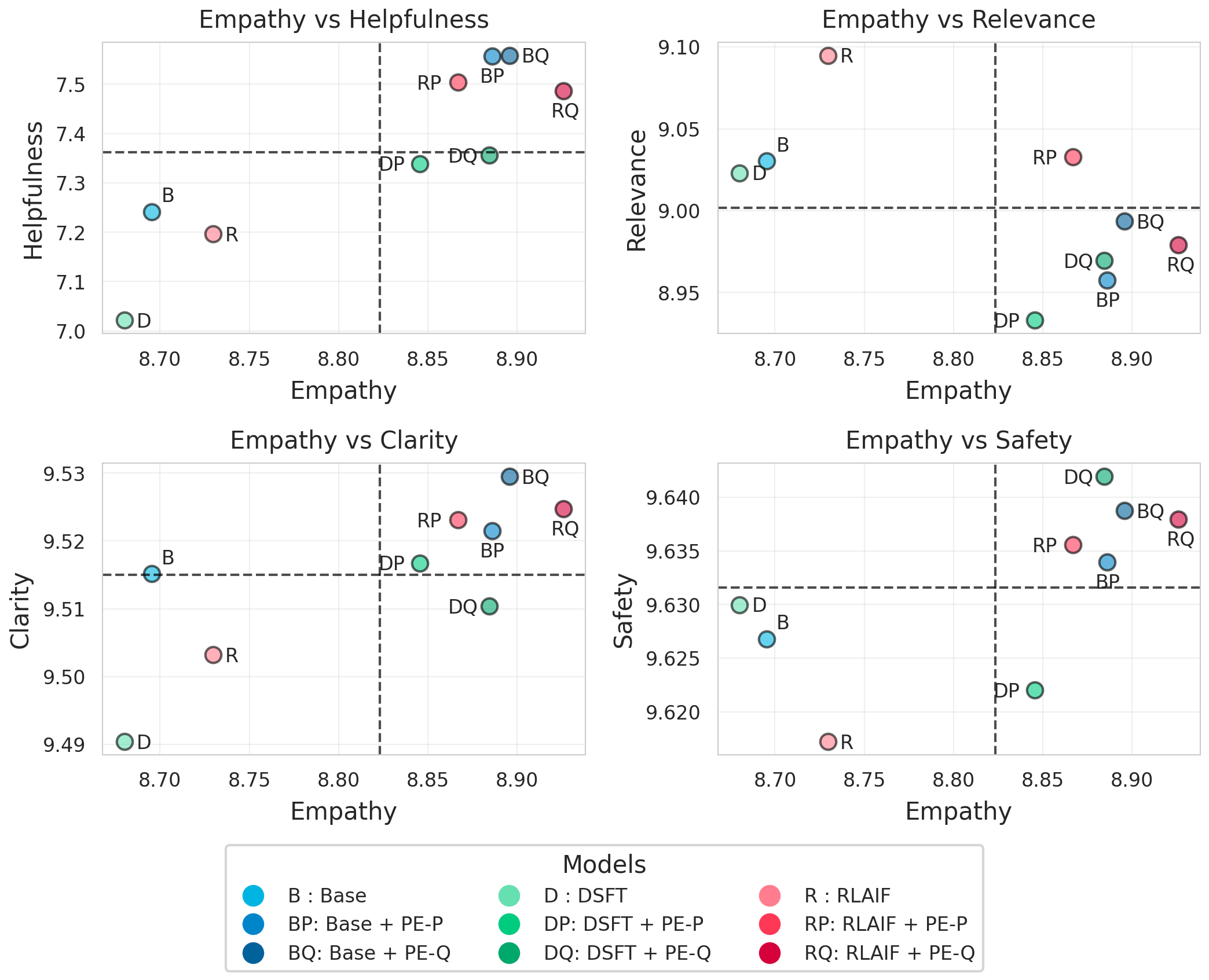}
    \caption{Empathy vs. Other Metrics Trade-offs Across Models and Prompting Strategies.}
    \label{fig:res_scatter}
\end{figure}

Third, the general-purpose Mistral model consistently outperformed the specialised MentalLLaMA model in all configurations (e.g., Base: 9.12 vs. 8.44), raising concerns about the effectiveness of specialised models for evaluations, as they may lack the generalisation capabilities needed for robust assessment.
These findings should be approached cautiously, as AI evaluation may overlook critical clinical factors like therapeutic empathy and emotional appropriateness, and the narrow scoring range (8.36–9.28) indicates a need for more sensitive assessment metrics in future research, including human expert evaluations.
Figure \ref{fig:res_scatter} presents scatter plots comparing empathy against four key quality dimensions: helpfulness, relevance, clarity, and safety, across experimental models. Each quadrant is divided by the median values of both axes to aid visual interpretation.

Models enhanced through prompt engineering (PE-P, PE-Q) consistently outperform others in both empathy and the secondary assessed dimension. The Base + PE-Q (BQ) and RLAIF + PE-Q (RQ) variants excel in all plots, indicating that qualitative prompt engineering enhances both empathy and overall response quality. The Empathy vs. Relevance plot shows PE models with slightly lower relevance scores, suggesting a trade-off where increased empathy may reduce relevance. The close clustering of BQ, RQ, and DQ in the Empathy vs Safety plot indicates improved emotional tone and reduced safety risks.

Overall, qualitative prompt engineering proves to be an effective strategy for enhancing empathetic and quality response, highlighting that prompting is more crucial than advanced fine-tuning alone in emotionally nuanced contexts.

\subsubsection{Perspective API Evaluation}

The Perspective API evaluation results (Table \ref{tab:result_perspective}) indicate that prompt engineering, especially the qualitative approach (PE-Q), significantly enhances the ethical and emotional quality of responses. Across all three models (Base, DSFT, and RLAIF), PE-Q reduced toxicity and incoherence while increasing compassion and respect. For example, the Base + PE-Q model achieved the lowest toxicity (0.0156) and highest compassion (0.9415).

\begin{table}[htb!]
\caption{Perspective API Evaluation of Mental Health Model}
\centering
\small
\begin{tblr}{Q[1.8cm,valign=t]Q[0.8cm,halign=r,valign=t]Q[0.95cm,halign=r,valign=t]Q[0.6cm,halign=r,valign=t]Q[0.6cm,halign=r,valign=t]Q[1.2cm,halign=r,valign=t]Q[0.6cm,halign=r,valign=t]Q[0.6cm,halign=r,valign=t]X[j,halign=r,valign=t]}
\hline
\tiny \textbf{Model} & \tiny \textbf{Toxicity} & \tiny \textbf{Identity Attack} & \tiny \textbf{Insult} & \tiny \textbf{Threat} & \tiny \textbf{Compassion} & \tiny \textbf{Respect} & \tiny \textbf{Nuance} & \tiny \textbf{Incoherent} \\ 
\hline
\tiny Base & \tiny 0.0166 & \tiny 0.0022 & \tiny 0.0094 & \tiny 0.0070 & \tiny 0.9250 & \tiny 0.8815 & \tiny 0.4387 & \tiny 0.1748 \\ 
\tiny Base + PE-Q & \tiny \textbf{0.0156} & \tiny \textbf{0.0021} & \tiny \textbf{0.0091} & \tiny \textbf{0.0069} & \tiny \textbf{0.9415} & \tiny \textbf{0.8838} & \tiny 0.4452 & \tiny \textbf{0.1506} \\ 
\tiny Base + PE-P & \tiny 0.0180 & \tiny 0.0022 & \tiny 0.0109 & \tiny 0.0070 & \tiny 0.9391 & \tiny 0.8814 & \tiny \textbf{0.4638} & \tiny 0.1564 \\ 
\hline
\tiny DSFT & \tiny \textbf{0.0174} & \tiny \textbf{0.0022} & \tiny \textbf{0.0096} & \tiny 0.0072 & \tiny 0.9246 & \tiny 0.8817 & \tiny 0.4092 & \tiny 0.1971 \\ 
\tiny DSFT + PE-Q & \tiny 0.0178 & \tiny 0.0023 & \tiny 0.0102 & \tiny \textbf{0.0069} & \tiny \textbf{0.9386} & \tiny \textbf{0.8841} & \tiny 0.4268 & \tiny \textbf{0.1690} \\ 
\tiny DSFT + PE-P & \tiny 0.0178 & \tiny 0.0023 & \tiny 0.0099 & \tiny 0.0071 & \tiny 0.9380 & \tiny 0.8813 & \tiny \textbf{0.4330} & \tiny 0.1747 \\ 
\hline
\tiny RLAIF & \tiny 0.0166 & \tiny \textbf{0.0021} & \tiny \textbf{0.0092} & \tiny 0.0072 & \tiny 0.9240 & \tiny 0.8782 & \tiny 0.4333 & \tiny 0.1772 \\ 
\tiny RLAIF + PE-Q & \tiny \textbf{0.0162} & \tiny 0.0022 & \tiny 0.0094 & \tiny \textbf{0.0069} & \tiny \textbf{0.9396} & \tiny \textbf{0.8810} & \tiny 0.4394 & \tiny \textbf{0.1442} \\ 
\tiny RLAIF + PE-P & \tiny 0.0171 & \tiny 0.0023 & \tiny 0.0097 & \tiny 0.0070 & \tiny 0.9375 & \tiny 0.8778 & \tiny \textbf{0.4527} & \tiny 0.1515 \\ 
\hline
\end{tblr}
\label{tab:result_perspective}
\end{table}

Though PE-P also improved outcomes, its effects were less consistent than PE-Q. Fine-tuning methods (DSFT and RLAIF) offered minimal ethical improvements without prompt enhancement, underscoring the need for well-designed prompts. These findings highlight the value of incorporating ethical safety metrics, particularly in sensitive fields like mental health. The Perspective API provides a scalable way to detect harmful tendencies and ensure alignment with compassionate communication standards in model evaluation.

\section{Conclusion and Future Work}
This study indicates that prompt engineering, particularly qualitative prompting, significantly enhances empathy and overall response quality in mental health chatbot applications. In contrast, advanced fine-tuning techniques such as DSFT and RLAIF offer only marginal improvements, suggesting that current fine-tuning strategies may not effectively capture the nuanced demands of emotionally sensitive dialogue. Although AI-based evaluation provides scalability and consistency, it remains limited in capturing clinically meaningful aspects such as emotional appropriateness, and long-term psychological impact. Future work will incorporate human evaluations from mental health professionals, and refine prompt generation using reinforcement learning from expert feedback.

%
%
%
\bibliographystyle{splncs04}
\bibliography{research_ref_short}

\end{document}